\documentclass[10pt,twocolumn,letterpaper]{article}

\usepackage{iccv}
\usepackage{times}
\usepackage{epsfig}
\usepackage{graphicx}
\usepackage{amsmath}
\usepackage{amssymb}
\usepackage{booktabs}
\usepackage{url}
\usepackage{epstopdf}
\usepackage{cite}
\usepackage{gensymb}
\usepackage{url}
\usepackage[bottom]{footmisc}
\usepackage{enumitem}
\usepackage{float}
\usepackage{subcaption}
\usepackage{multirow} 
\usepackage{color}
\allowdisplaybreaks
\usepackage{breqn}
\usepackage{tabularx}

\newcommand*{\Scale}[2][4]{\scalebox{#1}{$#2$}}%
\iccvfinalcopy 

\def\iccvPaperID{4889} 

\ificcvfinal\fi
\begin{document}

\title{Landmark Detection in Low Resolution Faces with Semi-Supervised Learning}

\author{  Amit Kumar \hspace{15pt} Rama Chellappa\\
Department of Electrical and Computer Engineering, CFAR and UMIACS\\
     University of Maryland-College Park,USA \\
{\tt\small akumar14@umiacs.umd.edu, rama@umiacs.umd.edu}
}

\maketitle

\begin{abstract}
Landmark detection algorithms trained on high resolution images perform poorly on datasets containing low resolution images. This deters the performance of algorithms relying on quality landmarks, for example, face recognition. To the best of our knowledge, there does not exist any dataset consisting of low resolution face images along with their annotated landmarks, making supervised training infeasible. In this paper, we present a semi-supervised approach to predict landmarks on low resolution images by learning them from labeled high resolution images. The objective of this work is to show that predicting landmarks directly on low resolution images is more effective than the current practice of aligning images after rescaling or super-resolution. In a two-step process, the proposed approach first learns to generate low resolution images by modeling the distribution of target low resolution images. In the second stage, the roles of generated images and real low resolution images are switched and the model learns to predict landmarks for real low resolution images from generated low resolution images. With extensive experimentation, we study the impact of each of the design choices and also show that prediction of landmarks directly on low resolution images improves the performance of important tasks such as face recognition in low resolution images.      
\end{abstract}

\vspace{-8pt}
\section{Introduction}
\label{intro}
Convolution Neural Networks (CNNs) have revolutionized the computer vision research, to the point that current systems can recognize faces with more than 99.7\%\cite{DBLP:journals/corr/abs-1801-07698} accuracy or achieve detection, segmentation and pose estimation results upto subpixel accuracy. These are only few of the many tasks which have seen a significant performance improvements in the last five years. However, CNN-based methods assume access to good quality images. ImageNet\cite{ILSVRC15}, COCO\cite{10.1007/978-3-319-10602-1_48}, CASIA\cite{DBLP:journals/corr/YiLLL14a}, 300W\cite{6755925} or MPII\cite{andriluka14cvpr} datasets all consist of high resolution images. As a result of \textit{domain shift} much lower performance is observed when networks trained on these datasets are applied to images which have suffered degradation due to intrinsic or extrinsic factors. In this work, we address landmark localization in low resolution images. Although, we use face images in our case, the proposed method is also applicable to other tasks, such as human pose estimation. Throughout this paper we use \textit{\textbf{HR and LR to denote high and low resolutions}} respectively. 
 
\begin{figure}[t]
\centering
\includegraphics[width=0.48\textwidth]{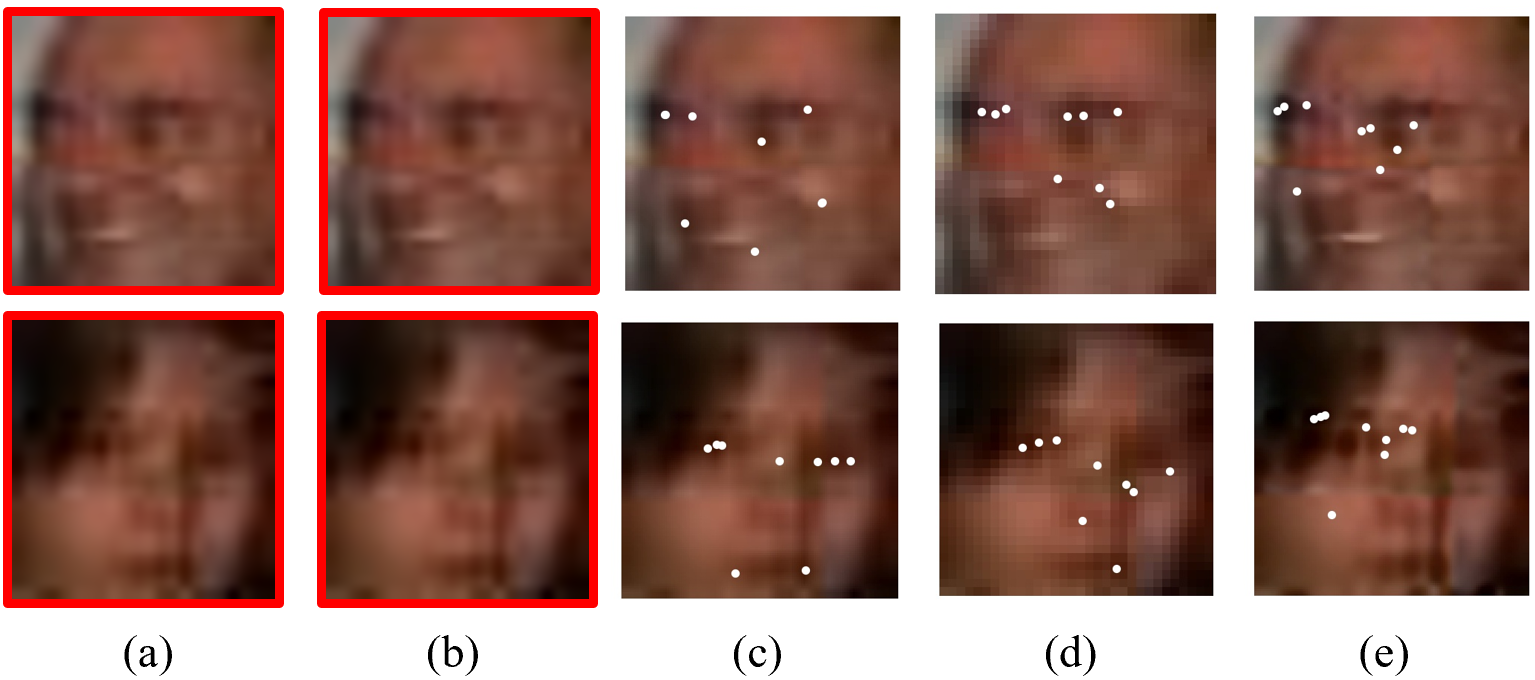}
\caption{\small Inaccurate landmark detections on low resolution images. We show landmark predicted by different systems. (a) MTCNN\cite{7553523} and (b) \cite{bulat2017far} are not able to detect any face in the LR image. (c) Current practice of directly upsampling the low-resolution image to a fixed size of $128\times128$ by bilinear interpolation. (d) Output from a network trained on downsampled version of HR images. (e) Landmark detection using super-resolved images. 
\textbf{Note:} For visualization purposes images have been reshaped after respective processing. Actual size of the images is in the range of $20\times20$ pixels}
\vspace{-15pt}
\label{fig:motivation}
\end{figure}

Facial landmark localization, also known as keypoint or fiducial detection, refers to the task of detecting specific points such as eye corners and nose tip on a face image. The detected keypoints are used to align images to canonical coordinates, which are then used as inputs to different convolution networks. It has been experimentally shown in \cite{bansal2017dosanddonts}, that accurate face alignment leads to improved performance in face verification. Though great strides have been made in this direction, mainly addressing large-pose face alignment, landmark localization for low resolution images, still remains an understudied problem, mostly because of the absence of large scale labeled dataset(s). To the best of our knowledge, for the first time, landmark localization directly on low resolution images is addressed in this work. 

\textbf{Main motivation:} In Figure \ref{fig:motivation}, we examine possible scenarios which are currently practiced when low resolution images are encountered. Figure \ref{fig:motivation} shows the predicted landmarks when the input image is a LR image of size less than $32\times32$ pixels. Typically, landmark detection networks are trained with $224\times224$ crops of HR images taken from AFLW\cite{tugraz:icg:lrs:koestinger11b} and 300W\cite{6755925} datasets. During inference, irrespective of resolution, an incoming image is rescaled to $224\times224$. We deploy two methods: MTCNN\cite{7553523} and Bulat \emph{et al.}\cite{bulat2017far}, which have detection and localization built in a single system. In Figure \ref{fig:motivation}(a) and (b) we see that these networks failed to detect face in the given image. Figure \ref{fig:motivation}(c), shows the outputs when a network trained on high resolution images is applied to a rescaled low resolution one. It is important to note that the trained network, say HR-LD high resolution landmark detector (detailed in Section \ref{additional}) achieves state of the art performance on AFLW and 300W test sets. A possible solution is to train a network on sub-sampled images as a substitute for low resolution images. Figure \ref{fig:motivation}(d) shows the output of one such network. It is evident from these experiments that networks trained with HR images or subsampled images are not effective for real life LR images. It can also be concluded that subsampled images are unable to capture the distribution of real LR images. 

Super-resolution is widely used to resolve LR images to reveal more details. Significant developments have been made in this field and methods based on encoder-decoder architectures and GANs\cite{NIPS2014_5423} have been proposed. We employ two recent deep learning based methods, SRGAN\cite{DBLP:journals/corr/LedigTHCATTWS16} and ESRGAN\cite{DBLP:journals/corr/abs-1809-00219} to resolve given LR images. It is worth noting that the training data for these networks also include face images. Figure \ref{fig:motivation}(e) shows the result when the super-resolved image is passed through HR-LD. It can be hypothesized that possibly, the super-resolved images do not lie in the same space of images using which HR-LD was trained. Super resolution networks are trained using synthetic low resolution images obtained by downsampling the image after applying Gaussian smoothing. In some cases, training data for super-resolution networks consists of paired low and high resolution images. Neither of the mentioned scenarios is applicable in real life situations.  

\textbf{Main Idea:} Different from these approaches, the proposed method is based on the concept of `generate to adapt'. This work aims to show that landmark localization in LR images can not only be achieved, but it also improves the performance over the current practice. To this end, we first train a deep network which generates LR images from HR images and tries to model the distribution of real LR images in pixel space. 
Since, there is no publicly available dataset, containing low resolution images along with landmark annotations, we take a semi-supervised approach for landmark detection. We train an adversarial landmark localization network on the generated LR images and hence, switching the roles of generated and real LR images. 
Heatmaps predicted for unlabelled LR images are also included in the inputs of the discriminators. The adversarial training procedure is designed in a way that in order to fool the discriminators, the heatmap generator has to learn the structure of the face even in low resolution. We perform extensive set of experiments explaining all the design choices. In addition, we also propose new state of the art landmark detector for HR images.  

\section{Related Work}

Being one of the most important pre-processing steps in face analysis tasks, facial landmark detection has been a topic of immense interest among computer vision researchers. We briefly discuss some of the methods which use Convolution Neural Networks (CNN). Different algorithms have been proposed in the recent past such as direct regression approaches of MTCNN by Zhang \emph{et al.} \cite{DBLP:conf/eccv/ZhangLLT14} and KEPLER by Kumar \emph{et al.} \cite{7961750}. The convolution neural networks in MTCNN and KEPLER act as non-linear regressors and learn to directly predict the landmarks. Both works are designed to predict other attributes along with keypoints such as 2D pose, visibility of keypoints, gender and many others. Hyperface by Ranjan \emph{et al.} \cite{DBLP:journals/corr/RanjanPC16}  has shown that learning tasks in one single network does in fact, improves the performance of individual tasks. Recently, architectures based on Encoder-Decoder architecture have become popular and have been used intensively in tasks which require per-pixel labeling such as semantic segmentation\cite{noh2015learning,Ronneberger2015UNetCN} and keypoint detection\cite{DBLP:journals/corr/abs-1802-06713,recdec,CFAN,DBLP:journals/corr/KumarC17a}. Despite making significant progress in this field, predicting landmarks on low resolution faces still remains a relatively unexplored topic. 
All of the works mentioned above are trained on high quality images and their performance degrades on LR images. 
\begin{figure*}[ht]
\centering
\includegraphics[height=5cm, width=0.8\textwidth]{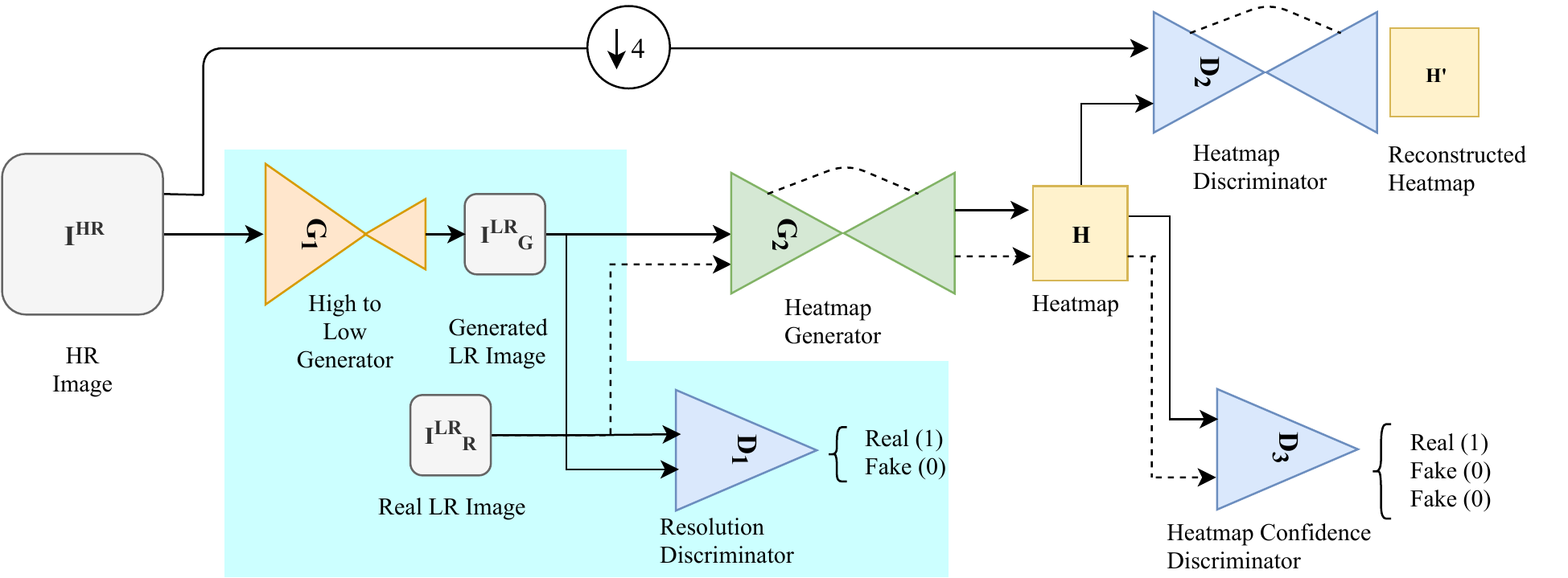}
\caption{\small Overview of the proposed approach. High resolution input is passed through High-to-Low generator $G_1$ (shown in cyan colored block). The discriminator $D_1$ learns to distinguish generated LR images vs. real LR images in an unpaired fashion. This generated image is fed to heatmap generator $G_2$. Heatmap discriminator $D_2$ distinguishes generated heatmap vs. groundtruth heatmaps. The pair $G_2, D_2$ is inspired from BEGAN\cite{DBLP:journals/corr/BerthelotSM17}. In addition to generated and groundtruth heatmaps, the discriminator $D_3$ also receives predicted heatmaps for real LR images. This enables the generator $G_2$ to generate realistic heatmaps for un-annotated LR images.}
\label{fig:flow}
\vspace{-8pt}
\end{figure*}
\vspace{-10pt}
\par One of the closely related works, is Super-FAN\cite{DBLP:journals/corr/abs-1712-02765} by Bulat \emph{et al.}, which makes an attempt to predict landmarks on LR images by super-resolution. However, as shown in experiments in Section \ref{experiments}, face recognition performance degrades even on super-resolved images. This necessitates that super-resolution, face-alignment and face recognition be learned in a single model, trained end to end, making it not only slow in inference but also limited by the GPU memory constraints. The proposed work is different from \cite{DBLP:journals/corr/abs-1712-02765} in many respects as it needs labeled data only in HR and learns to predict landmarks in LR images in an unsupervised way. Due to adversarial training, the network not only acts as a facial parts detector but also learns the inherent structure of the facial parts. The proposed method makes the pre-processing task faster and independent of face verification training. During inference, only the heatmap generator network is used which is based on the fully convolutional architecture of U-Net\cite{Ronneberger2015UNetCN} and works at the spatial resolution of $32\times32$ making the alignment process real time. 
\vspace{-5pt}
\section{Proposed Method}
\label{proposed}
The proposed work predicts landmarks directly on a low resolution image of spatial size less than $32\times32$ pixels. We show that predicting landmark detection directly in low resolution is effective than current practices of rescaling or super-resolution. The entire pipeline can be divided into two stages: (a) Generation of LR images in an unpaired manner (b) Generating heatmaps for real LR images in a semi-supervised fashion. The diagrammatic overview of the proposed approach is shown in Figure \ref{fig:flow}. Being a semi-supervised method, it is important to first describe the datasets chosen for the ablative study. 

\textbf{High Resolution Dataset:} We construct the HR dataset by combining the $20,000$ training images from AFLW\cite{tugraz:icg:lrs:koestinger11b} and the entire 300W\cite{6755925} dataset. We divide the Widerface dataset\cite{yang2016wider} which consists of images in different resolutions captured under diverse conditions, into two groups based on their spatial size. The first group consists of images with spatial size between $20\times20$ and $40\times40$, whereas the second group consists of images with more than $100\times100$ pixels. We combine the second group in HR training set, resulting in a total of $35,543$ HR faces. The remaining $4,386$ images from AFLW are used as validation images for the ablative study and test set for the landmark localization task. Although, generation of LR images is an unpaired task, we use AFLW and 300W images for training, as the generated LR images from these datasets are used for semi-supervised learning in the second step. 

\textbf{Low Resolution Dataset}: The first group from Widerface dataset consists of $47,046$ faces and is used as real or target low resolution images.

\subsection{High to Low Generator and Discriminator }
\begin{figure}[h]
\centering
\includegraphics[width=0.5\textwidth]{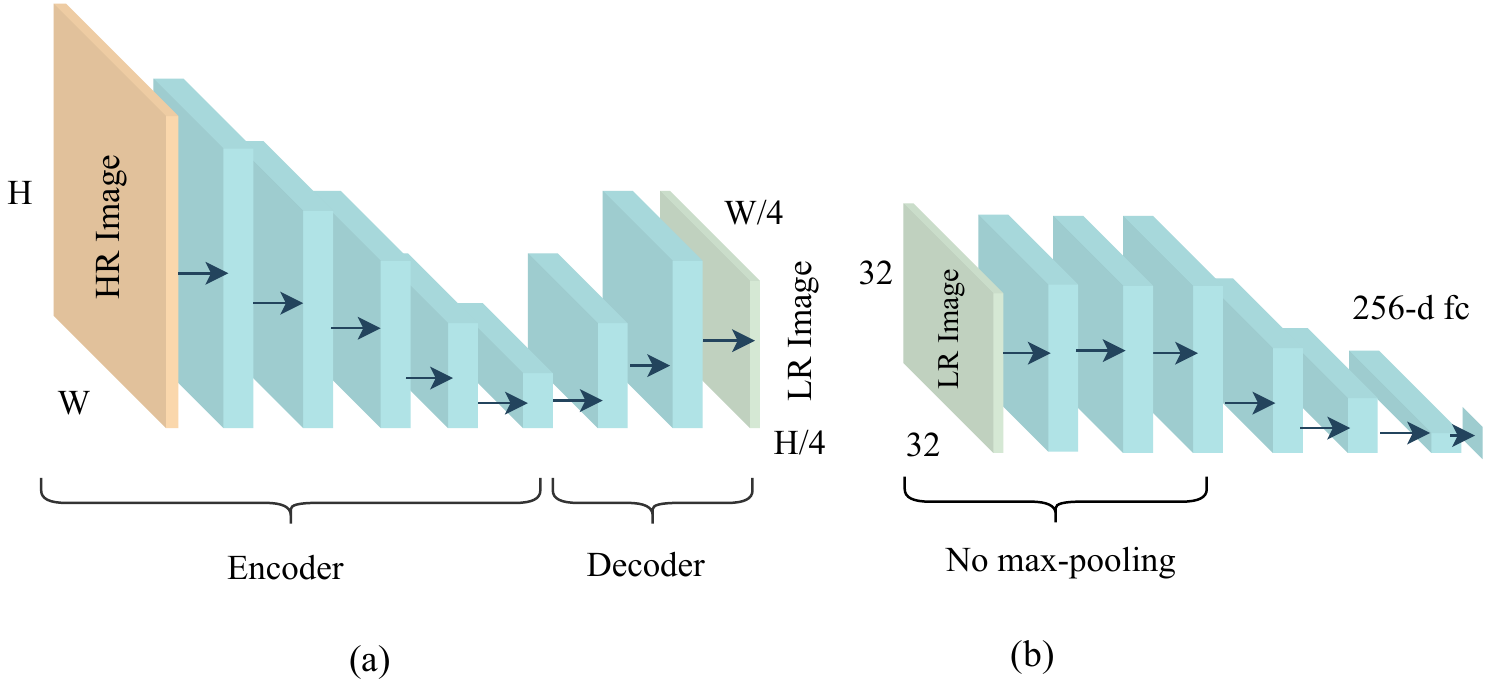}
\caption{\small (a) Generator used in high to low resolution generator $G_1$. Each $\rightarrow$ represents two residual blocks followed by a convolution layer. (b) Discriminator used in $D_1$ and $D_2$. Max-pooling is applied only in the last two layers. Each $\rightarrow$ represents one residual block followed by a convolution layer.}
\label{fig:hightolow}
\vspace{-5pt}
\end{figure}

High to low generator $G_1$, shown in Figure \ref{fig:hightolow} is designed following the Encoder-Decoder architecture, where both encoder and decoder consists of multiple residual blocks. The input to the first convolution layer is the HR image concatenated with the noise vector which has been projected using a fully connected layer and reshaped to match the input size. Similar architectures have also been used in \cite{bulatyang2018learn,DBLP:journals/corr/LedigTHCATTWS16}. The encoder in the generator consists of eight residual blocks each followed by a convolution layer to increase dimensionality. Max-pooling is used to decrease the spatial resolution to $4\times4$, for high resolution image of $128\times128$ pixels. The decoder is composed of six residual units followed by convolution layers to reduce the dimensionality. Finally, one convolution layer is added in order to output a three channel image.  
BatchNorm is used after every convolution layer. 

The discriminator $D_1$, shown in Figure \ref{fig:hightolow} is also constructed in a similar way, except max-pooling is used only in the last three layers considering the inputs to discriminator are low resolution $32\times32$ images. Referring to Figure \ref{fig:flow}, we use $I^{HR}$ for input high resolution images of size $128\times128$, $I^{LR}_{G}$ for generated LR images of size $32\times32$ and $I^{LR}_{R}$ for real LR images of the same size. 


We train High to Low generator using a weighted combination of GAN loss and $L_{2}$ pixel losses. $L_{2}$ loss is used to encourage convergence in initial training iterations. The final loss can be summarized in Equation \ref{totalloss}. 
\begin{equation}
\label{totalloss}
\Scale[0.9]{l = \alpha l_{GAN} + \beta l_{pixel}}
\end{equation}
where $\alpha$ and $\beta$ are hyperpameters which are empirically set following $\alpha l_{GAN}>\beta l_{pixel}$. Following recent developments in GANs we experimented with different loss functions. However, we use Hinge loss and Spectral Normalization\cite{DBLP:journals/corr/abs-1802-05957} in combination due to faster training. The hinge loss for the generative networks can be defined as in Equation \ref{ganloss}:
\begin{equation}
\label{ganloss}
\Scale[0.85]{l_{GAN} = \mathop{\mathbb{E}}_{x\in\mathbb{P}_r}[\mathrm{min}(0, -1+D_{1}(x))] + \mathop{\mathbb{E}}_{\hat{x}\in\mathbb{P}_g}[\mathrm{min}(0,-1-D_{1}(\hat{x}))]}
\end{equation}
where $\mathbb{P}_r$ is the distribution of real LR images $I^{LR}_{R}$ from Widerface dataset, and $\mathbb{P}_g$ is the distribution of generated images $I^{LR}_{G}$. 

The weights of the discriminator $D_{1}$ are normalized in order to satisfy the Lipschitz constraint $\sigma(W)=1$, shown in Equation \ref{sn}:
\begin{equation}
\label{sn}
\Scale[0.9]{W_{SN}(W) = \frac{W}{\sigma(W)}}
\end{equation}
Finally, $L_{2}$ pixel loss described in Equation \ref{l2loss} is used which minimizes the distance between the generated and subsampled images. The $L_{2}$ loss ensures that the content is not lost during the generation process.
\begin{equation}
\label{l2loss}
\Scale[0.9]{l_{pixel} = \frac{1}{WH}\sum\limits_{i=1}^{W}\sum\limits_{i=1}^{H}(F(I^{HR})-I^{LR}_{G})^2}
\end{equation}   
where the operation $F$ is implemented as a sub-sampling operation obtained by passing $I^{HR}$ through four average pooling layers.
\begin{figure}[t]
\centering
\includegraphics[height=4cm, width=0.48\textwidth]{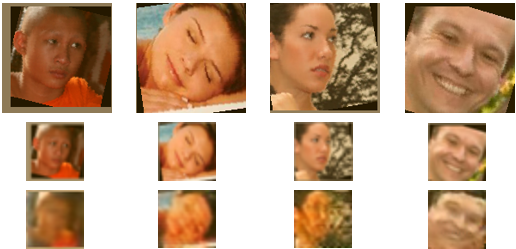}
\caption{\small Sample outputs of High to Low generator. First row shows the HR images. Second row shows downsampled images obtained after applying Gaussian smoothing. Third row shows LR images generated by the network. \textbf{Note:} Best viewed when zoomed in. For visualization purposes, images have been enlarged after respective processing. Actual size of the images is in the range of $20\times20$ pixels}
\vspace{-5pt}
\label{fig:genlowres}
\end{figure}
Figure \ref{fig:genlowres} shows some sample LR images generated from the network $G_1$.
\subsection{Semi-Supervised Landmark Localization}

\subsubsection{Heatmap Generator $G_2$} 
\begin{figure}[h]
\centering
\includegraphics[height=3.5cm,width=0.35\textwidth]{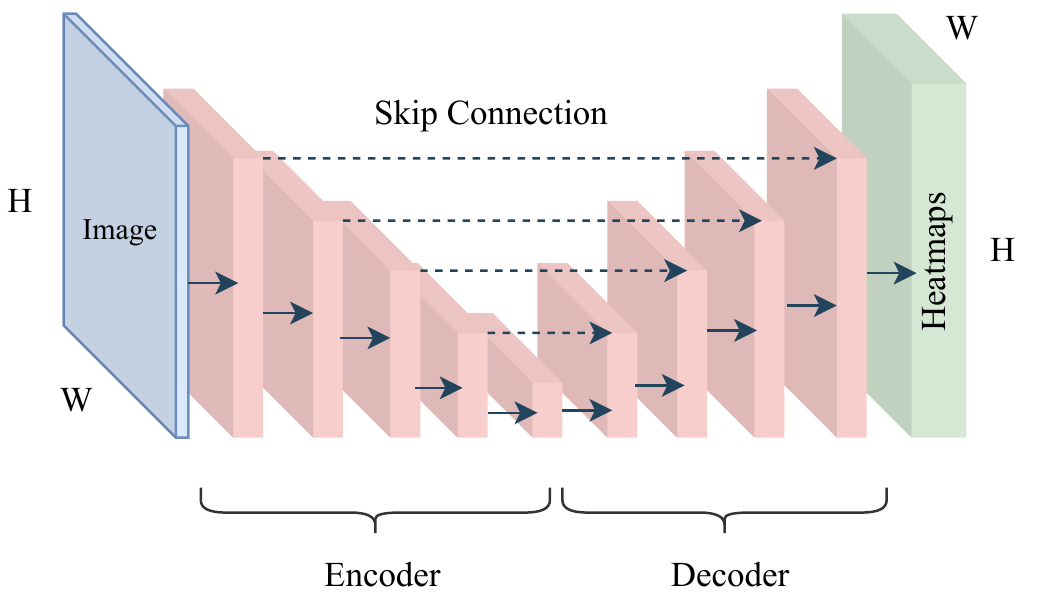}
\caption{\small Architecture of the heatmap generator $G_2$. Architecture of this network is based on U-Net. Each $\rightarrow$ represents two residual blocks. $\dashrightarrow$ represents skip connections between the encoder and decoder. }
\label{fig:unet}
\vspace{-5pt}
\end{figure}
The keypoint heatmap generator, $G_2$ in Figure \ref{fig:unet} produces heatmaps corresponding to N (in our case $19$ or $68$) keypoints in a given image. As mentioned earlier, the objective of this paper is to show that landmark prediction directly on LR image is feasible even in the absence of labeled LR data, and evaluate the performance of auxiliary tasks compared to commonly used practices of rescaling or super-resolution. Keeping this in mind, we choose a simple network based on the U-Net\cite{Ronneberger2015UNetCN} architecture as the heatmap generator, instead of computationally intensive stacks of hourglass networks\cite{Newell2016} or CPMs\cite{DBLP:journals/corr/WeiRKS16}. The network consists of 16 residual blocks where both encoder and decoder have eight residual blocks. Eight residual blocks in the encoder are divided into four groups of two blocks each and spatial resolution is halved after each block using max pooling. The heatmap generator outputs (N+1) feature maps corresponding to N keypoints and 1 background channel. After experimentation, this design for landmark detection has proven to be very effective and has resulted in state of the art results for landmark predictions when trained with HR images (see Section \ref{experiments}).
\vspace{-5pt}
\subsubsection{Heatmap Discriminator $D_2$} The heatmap discriminator $D_2$ follows the same architecture as the heatmap generator. However, the input to the discriminator is a set of heatmaps concatenated with their respective color images. This discriminator predicts another set of heatmaps and learns whether the keypoints described by the heatmaps are correct and correspond to the face in the input image. The qualities of the output heatmaps are determined by their similarity to the input heatmaps, following the notion of an autoencoder. The loss is computed as the error between the input and the reconstructed heatmaps.
\vspace{-5pt}
\subsubsection{Heatmap Confidence Discriminator $D_3$} The architecture of heatmap confidence discriminator $D_3$ is identical to the one used in high to low discriminator, except the input is an LR image concatenated with the heatmap. This discriminator receives three inputs corresponding to the generated LR image with groundtruth heatmap, generated LR image with predicted heatmap and a real LR image with predicted heatmap. This discriminator learns to distinguish between the groundtruth and predicted heatmaps. In order to fool this discriminator, the generator should generate heatmaps which are as real or feasible (for unlabeled real LR image) as possible. The loss propagated from this discriminator enforces the generator to learn, not only to predict accurate heatmaps for images whose groundtruth are available but also for the images without annotations. This in turn enables the generator to understand the structure of the face in the given image and make accurate predictions. 

\textbf{Switching roles of generated and real images:} During training of this part of the system, the roles of generated and low resolution images are switched. While training High to Low discriminator $D_{1}$, the generated LR images are considered to be fake so that the generator tries to generate as realistic LR image as possible. It is worth recalling that HR images have annotations associated with them. We assume that keypoint locations in a generated LR image stay relatively same as its downsampled version. Therefore, while training $G_2$, the downsampled annotations are considered to be groundtruth for the generated LR images, and the networks are trained to predict heatmaps as close to the groundtruth as possible in order to fool the discriminator $D_{2}$ and $D_3$. $G_2$ tries to predict accurate keypoints for real LR images by learning from generated LR images, and hence the switching of roles. 

\subsection{Semi-supervised Learning}
The learning process of this setup is inspired by the seminal work of Berthelot \emph{et al.} in \cite{DBLP:journals/corr/BerthelotSM17} and Lecun \emph{et al.} in \cite{DBLP:journals/corr/ZhaoML16} called Energy-based GANs. The discriminator $D_{2}$ receives two sets of inputs: generated LR image with downsampled groundtruth heatmaps and generated LR images with predicted heatmaps. When the input consists of groundtruth heatmaps, the discriminator is trained to recognize it and reconstruct a similar one, i.e., to minimize the error between the groundtruth heatmaps and the reconstructed ones. On the other hand, if the input consists of generated heatmaps, the discriminator is trained to reconstruct different heatmaps, i.e., to drive the error between the generated heatmaps and the reconstructed heatmaps as large as possible. The losses are expressed as
\begin{equation}
\label{keydisc1}
\Scale[0.9]{l_{D}^{real} = \sum\limits_{i=1}^{N+1}(H_{i}-D_{2}(H_{i},I))^2  }
\end{equation}
\begin{equation}
\label{keydisc2}
\Scale[0.9]{l_{D}^{fake} = \sum\limits_{i=1}^{N+1}(\hat{H}_{i}-D_{2}(\hat{H}_{i},I))^2 }
\end{equation}
\begin{equation}
\label{keydisc3}
\Scale[0.9]{l_{D}^{kp} = l_{D}^{real} - k_{t}l_{D}^{fake}}
\end{equation}
where $H_{i}$ represents the $i^{th}$ heatmap of a given image $I$ constructed by placing Gaussian with $\sigma=2$ centered at the keypoint location $(x_{i},y_{i})$. Inspired by Berthelot et.al. in \cite{DBLP:journals/corr/BerthelotSM17}, we use
a variable $k_{t}$ to control the balance between heatmap generator and discriminator. The variable is updated every $t$ iterations. The adaptive term $k_{t}$ is defined by:
\begin{equation}
\label{discupdate}
\Scale[0.9]{k_{t+1} = k_{t} + \lambda_{k}(\gamma l_{D}^{real}-l_{D}^{fake})}
\end{equation}
where $k_{t}$ is bounded between $0$ and $1$, and $\lambda_{k}$ is a hyperparameter. As in Equation \ref{keydisc3}, $k_{t}$ controls the emphasis on $l_{D}^{fake}$. When the generator is able to fool the discriminator, $l_{D}^{fake}$ becomes smaller than $\gamma l_{D}^{real}$. As a result of this $k_{t}$ increases, making the term $l_{D}^{fake}$ dominant. The amount of acceleration to train on $l_{D}^{fake}$ is adjusted proportional to $\gamma l_{D}^{real}-l_{D}^{fake}$, \emph{i.e} the distance the discriminator falls behind the generator. Similarly, when the discriminator gets better than the generator, $k_{t}$ decreases, to slow down the training on $l_{D}^{fake}$ making the generator and the discriminator train together.

The discriminator $D_{3}$ is trained using the loss function from Least squares GAN\cite{DBLP:journals/corr/MaoLXLW16} as shown in Equation \ref{confdiscloss}. This loss function was chosen in order to be consistent with the losses computed by $D_2$ which are also $L_2$ losses. 
\begin{equation}
\label{confdiscloss}
\Scale[0.8]{l_{D}^{conf} = \mathop{\mathbb{E}}_{x\in\mathbb{P}_r}[(D_{3}(x)-1)^2] + \mathop{\mathbb{E}}_{\hat{x}\in\mathbb{P}_g}[D_{3}(\hat{x})^2] + \mathop{\mathbb{E}}_{\hat{y}\in\mathbb{P}_g}[D_{3}(\hat{y})^2]}
\end{equation}
It is noteworthy to mention that in this case $\mathbb{P}_r$ represents the groundtruth-heatmaps distribution on generated LR images, while $\mathbb{P}_g$ represents the distribution on generated heatmaps of generated LR images and real LR images.

\begin{figure}[t]
\centering
\includegraphics[width=0.48\textwidth]{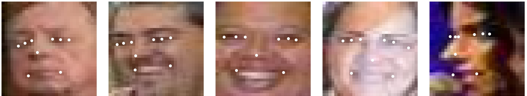}
\caption{\small Sample keypoint detections on Tinyface images. \textbf{Note:} For visualization purposes images have been enlarged after rocessing. Actual size of the images is in the range of $20\times20$ pixels. }
\label{fig:result_img}
\vspace{-10pt}
\end{figure}

The generator $G_{2}$ is trained using a weighted combination of losses from the discriminators $D_2$ and $D_3$ and $l_{MSE}$ heatmap loss. The loss functions for the generator $G_2$ are described in the following equations: 
\begin{equation}
\label{keygen1}
\Scale[0.9]{l_{G}^{MSE} =  \sum\limits_{i=1}^{N+1}(H_{i}-G_{2}(H_{i},I))^2}
\end{equation}
\begin{equation}
\label{keygen2}
\Scale[0.9]{l_{G}^{kp} = \sum\limits_{i=1}^{N+1}(\hat{H}_{i}-D_{2}(\hat{H}_{i},I))^2 }
\end{equation}
\begin{equation}
\label{keygen3}
\Scale[0.9]{l_{G}^{conf} = \mathop{\mathbb{E}}_{x\in\mathbb{P}_g}[(D_{3}(x)-1)^2]}
\end{equation}
\begin{equation}
\label{keygen4}
\Scale[0.9]{l_{G} = a l_{G}^{MSE}+ b l_{G}^{kp} + c l_{G}^{conf}}
\end{equation}
where $a, b$ and $c$ are hyper parameters set empirically obeying $a l_{G}^{MSE}>b l_{G}^{kp}>c l_{G}^{conf}$. We put more emphasis on $l_{G}^{MSE}$ to encourage convergence of the model in initial iterations. 
Some real LR images with keypoints predicted from the $G_2$ are shown in Figure \ref{fig:result_img}.
\section{Experiments and Results}
\subsection{Ablation Experiments}
\label{ablation}
We experimentally demonstrated in Section \ref{intro} (Figure \ref{fig:motivation}) that networks trained on HR images perform poorly on LR images. Therefore, we propose the semi-supervised learning as mentioned in Section \ref{proposed}. With the above mentioned networks and loss functions it is important to understand the implication of each component. This section examines each of the design choices quantitatively. To this end, we first train the high to low resolution networks, and generate LR images of $4,386$ AFLW test images. In the absence of real LR images with annotated landmarks, this is done to create a substitute for low resolution dataset with annotations on which localization performance can be evaluated. We also generate subsampled version of the $20,000$ AFLW trainset and $4,386$ AFLW testset using average pooling after applying Gaussian smoothing. Data augmentation techniques such as random scaling $(0.9,1.1)$, random rotation ($-30\degree,30\degree$) and random translation upto $20$ pixels are used. 

\textbf{Evaluation Metric:} Following most previous works, we obtain error for each test sample by averaging normalized errors for all annotated landmarks. For AFLW, the obtained error is normalized by the ground truth bounding box size over all visible points whereas for 300W, the error is normalized by the inter-pupil distance. Wherever applicable NRMSE stands for Normalized Root Mean Square Error. \vspace{2pt}

\textbf{Training Details:} All the networks are trained in Pytorch using the Adam optimizer with an initial learning rate of $2\mathrm{E}-4$ and $\beta_1, \beta_2$ values of $0.5, 0.9$. We train the networks with a batch size of $32$ for $200$ epochs, while dropping the learning rates by $0.5$ after $80$ and $160$ epochs. \vspace{2pt}

\textbf{Setting S1:} \textit{Train networks on subsampled images?} We only train network $G_{2}$ with the subsampled AFLW training images using the loss function in Equation \ref{keygen1}, and evaluate the performance on generated LR AFLW test images.  \vspace{2pt}

\textbf{Setting S2:} \textit{Train networks on generated LR images?} In this experiment, we train the network $G_{2}$ using generated LR images, in a supervised way using the loss function from Equation \ref{keygen1}. We again evaluate the performance on generated LR AFLW test images.

\textbf{\textit{Observation:}} From the results summarized in Table \ref{ablation-table} it is evident that there is a significant reduction in localization error when $G_2$ is trained on generated LR images validating our hypothesis that subsampled images on which many super-resolution networks are trained may not be a correct representative of real LR images. Hence, we need to train the networks on real LR images.   \vspace{2pt}

\textbf{Setting S3:} \textit{Does adversarial training help?} This question is asked in order to understand the importance of training the heatmap generator $G_2$ in an adversarial way. In this experiment, we train $G_{2}$ and $D_{2}$ using the losses in Eqs \ref{keydisc1}, \ref{keydisc2}, \ref{keygen1}, \ref{keygen2}. Metrics are calculated on the generated LR AFLW test images and compared against the experimental setting mentioned in S2 above.  \vspace{2pt}

\textbf{Setting S4:} \textit{Does $G_{2}$ trained in adversarial manner scale to real LR images?} In this experiment, we wish to examine if training networks $G_2, D_2$ and $D_3$ jointly, improves the performance on real LR images from Widerface dataset.(see Section \ref{proposed} for datasets) 

\textbf{\textit{Observation:}} From Table \ref{ablation-table} we observe that the network trained with setting S3 performs marginally better compared to setting S4. However, since there are no keypoint annotations available for the Widerface dataset, conclusions cannot be drawn from the drop in performance. Hence, in the following subsection \ref{experiments}, we leap towards understanding this phenomenon indirectly, by aligning the faces using the models from setting S3 and setting S4 and evaluating face recognition performances.

\begin{table}[t]
\begin{subtable}[h]{0.5\textwidth}
\centering
\footnotesize
\begin{tabular}{|c|ccc|} \hline
Method & NRMSE (all) & NRMSE (479 images) & Time \\ \hline
MTCNN\cite{7553523} & - & 0.9736 & 0.388 s\\
HRNet\cite{SunXLW19} & 0.4055 & 0.3107 & 0.076 s\\
SAN\cite{dong2018san} & 0.3901 & 0.3141 & 0.0178 s\\ \hline
\textbf{Proposed} & \textbf{0.257} & \textbf{0.1803} & \textbf{0.0105 s}\\ \hline 
\end{tabular}
  \caption{}
  \label{reallowresexp}
 \end{subtable}
\begin{subtable}[h]{0.5\textwidth}
\centering
\begin{tabular}{@{}|llll|@{}}
\hline
Setting   & NRMSE$\pm$std & auc@0.07 & auc@0.08 \\ \hline \hline
S1 & $11.33\pm 9.81$          &   11.897       &    21.894      \\
S2 & $4.23\pm4.52 $          &   50.843       &      55.751    \\
S3 &  $4.120 \pm 4.43$         &  51.889        &  56.791        \\
S4 &  $4.123 \pm 4.394$         &   51.775       &  56.697        \\ \hline
\end{tabular}
\caption{}
\label{ablation-table}
\end{subtable}
\caption{(a) Landmark Detection Error on Real Low Resolution dataset. (b) Table for ablation experiments under different settings on synthesized LR images.}
\end{table}
\subsection{Experiments on Low Resolution images}
We choose to perform direct comparison on a real LR dataset. Two recent state of the art methods Style Aggregated Networks\cite{dong2018san} and HRNet\cite{SunXLW19}. To create a real LR landmark detection dataset which we call Annotated LR Faces (ALRF), we randomly selected $700$ identities from the TinyFace dataset, out of which one LR image (less than $32\times 32$ pixels and more than $15\times 15$ pixels) per identity was randomly selected, resulting in a total of $700$ LR images. Next, three individuals were asked to manually annotated all the images with 5 landmarks(two eye centers, nose tip and mouth corners) in MTCNN\cite{7553523} style, where invisible points were annotated with $-1$. The mean of the points obtained from the three users were taken to be the groundtruth. As per convention, we used Normalised Mean Square Error (NRMSE), averaged over all visible points and normalized by the face size as the comparison metric. Table \ref{reallowresexp} shows the results of this experiment. We also calculate time for forward pass of one image in a single gtx1080. Without loss of generality, the results can be extrapolated to other existing works as \cite{dong2018san} and \cite{SunXLW19} are currently state of the art. MTCNN which has detection and alignment in a single system was able to detect only $479$ faces out of $700$ test images. 

\begin{figure}
\centering
  \fbox{\includegraphics[width = 0.4\textwidth]{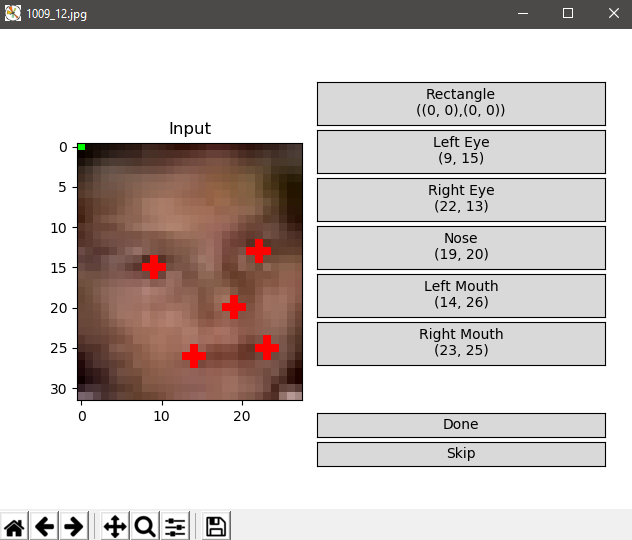}}%
{%
  \caption{\small Snippet of the annotation tool used.}%
}
\end{figure}
\subsection{Face Recognition experiments}
\label{experiments}
In the previous section, we performed ablative studies on the generated LR AFLW images. Although convenient to quantify the performance, it does not uncover the importance of training three networks jointly in a semi-supervised way. Therefore, in this section, we choose to evaluate the models from setting S3 and setting S4 (Section \ref{ablation}), by comparing the statistics obtained by applying the two models to align face images for face recognition task.

We use recently published and publicly available, Tinyface\cite{DBLP:journals/corr/abs-1811-08965} dataset for our experimental evaluation. It is one of the very few datasets aimed towards understanding LR face recognition and consists of $5,139$ labeled facial identities with an average of three face images per identity, giving a total of $15,975$ LR face images (average $20\times16$ pixels). All the LR faces in TinyFace are collected from the web (PIPA\cite{DBLP:journals/corr/ZhangPTFB15} and MegaFace2\cite{DBLP:journals/corr/NechK17}) across diverse imaging scenarios, captured under uncontrolled viewing conditions in pose, illumination, occlusion and background. $5,139$ known identities is divided into two splits: $2,570$ for training and the remaining $2,569$ for test.

\textbf{Evaluation Protocol:} In order to compare model performances, we adopt the closed-set face identification (1:N matching) protocol. Specifically, the task is to match a given probe face against a gallery set of enrolled face images with true match from the gallery at top-1 of the ranking list. For each test class, half of the face images are randomly assigned to the probe set, and the remaining to the gallery set. For the purpose of this paper, we drop the distractor set as this does not divulge new information while significantly slowing down the evaluation process. For face recognition evaluation, we report statistics on Top-k (k=1,5,10,20) statistics and mean average precision (mAP). 

\textbf{Experiments with network trained from scratch:} Since the number of images in TinyFace dataset is much smaller compared to larger datasets such as CASIA\cite{DBLP:journals/corr/YiLLL14a} or MsCeleb-1M\cite{DBLP:journals/corr/GuoZHHG16}, we observed that training a very deep model like Inception-ResNet\cite{DBLP:journals/corr/SzegedyIV16}, quickly leads to over-fitting. Therefore, we adopt a CNN with fewer parameters, specifically, LightCNN\cite{DBLP:journals/corr/WuHS15}. Since inputs to the network are images of size $32\times32$, we disable first two max-pooling layers. After detecting the landmarks, training and testing images are aligned to the canonical coordinates using affine transformation. We train $29$ layer LightCNN models using the training split of TinyFace dataset under the following settings: 

\begin{table}[]

\begin{subtable}[h]{0.5\textwidth}
\centering
\begin{tabular}{@{}|l|lllll|@{}}
\hline
Setting            &  L1 &  L2 &  L3 &  L4 & L5 \\ \hline \hline
top-1      &  31.17     &   35.11    &  39.03 & 39.87 & 43.82    \\ \hline
\end{tabular}
\caption{}
\label{tinyface-rank1}
\end{subtable}
\begin{subtable}[h]{0.5\textwidth}
\centering
\resizebox{\columnwidth}{!}{
\begin{tabular}{|l|lllll|}
\hline
Setting    & top-1 & top-5 & top-10 & top-20 & mAP \\ 
\hline \hline
Baseline (ArcFace\cite{DBLP:journals/corr/abs-1801-07698}) &    34.71   &   44.82  & 49.01 & 53.70 &  0.32   \\
I1 &    34.01   &   41.98    & 45.36 & 49.22 &  0.29   \\
I2 &   45.04    &   56.30    & 60.11 & 63.71 &  0.43   \\ \hline \hline
I3 &   \textbf{51.10}    &   \textbf{61.05}    & \textbf{64.38} & \textbf{67.89} &  \textbf{0.47}   \\ 
\hline
\end{tabular}
}
\caption{}
\label{tinyface_ir}
\end{subtable}
\vspace{-2pt}
\caption{\small Verification performance on Tinyface dataset under different settings (a) LightCNN trained from scratch (b) Using Inception-ResNet pretrained on MsCeleb-1M}
\label{tinyface-verification}
\vspace{-10pt}
\end{table}

\textbf{Setting L1:} \textit{Train networks on generated LR images?} In this setting, we use the model trained under the setting S2 from the previous section \ref{ablation}. In this setting, network $G_2$ is trained using generated LR images in a supervised way using the loss function from Equation \ref{keygen1}.  \vspace{2pt}

\textbf{Setting L2:} \textit{Does adversarial training help?} We use the model trained from setting S3 (section \ref{ablation}) to align the faces in training and testing sets. In this setting networks $G_2$ and $D_2$ are trained using a weighted combination of $L_2$ pixel loss and GAN losses from Equations \ref{keydisc1}, \ref{keydisc2}, \ref{keygen1}, \ref{keygen2}.  \vspace{2pt}

\textbf{Setting L3:} \textit{Does $G_{2}$ trained in adversarial manner scale to real LR images?} In this setting, networks $G_2$, $D_2$ and $D_3$ are trained jointly in a semi-supervised way. We use Tinyface training images as real low resolution images. Later, Tinyface training and testing images are aligned using the trained model for training LightCNN model. \vspace{2pt}

\textbf{Setting L4:} \textit{End-to-end training?} Under this setting, we also train the High to Low networks $G_1$ and $D_1$, using the training images from Tinyface dataset as real LR images. We reduce the amount of data-augmentation in this case to resemble tiny face dataset images. With the obtained trained model, landmarks are extracted and images are aligned for LightCNN training. \vspace{2pt}
 
\textbf{Setting L5:} \textit{End-to-end training with pre-trained weights?} This setting is similar to the setting L4 above, except instead of training a LightCNN model from scratch we initialize the weights from a pre-trained model, trained with CASIA-Webface dataset.

\textbf{\textit{Observation:}} The results in Table \ref{tinyface-rank1} summarizes the results of the experiments done under the settings discussed above. We see that although, we observed a drop in performance in landmark localization when training the three networks jointly (Table \ref{ablation-table}), there is a significant gap in rank-1 performance between setting L2 and L3. This indicates that with semi-supervised learning $G_2$ generalizes well to real LR data, and hence also validates our hypothesis of training $G_2$, $D_2$ and $D_3$ together. Unsurprisingly, insignificant difference is seen between settings L3 and L4. 

\textbf{Experiments with pre-trained network:} Next, to further understand the implications of joint semi-supervised learning, we design another set of experiments. In these experiments, we use a pre-trained Inception-ResNet model, trained on MsCeleb-1M using ArcFace\cite{DBLP:journals/corr/abs-1801-07698} and Focal Loss\cite{DBLP:journals/corr/abs-1708-02002}. This model expects an input of size $112\times112$ pixels, hence the images are resized after alignment in low resolution. Using this pre-trained network, we perform the following experiments: 
\begin{table}[h]
\centering
\begin{tabular}{@{}|l|lllll|@{}}
\hline
Setting           & top-1 & top-5 & top-10 & top-20 & mAP \\ \hline \hline
 A1 &   11.75    &   14.58  & 24.57 & 30.47 &  0.10   \\
 A2 &   26.21   &   34.76  & 39.03 & 43.99 &  0.24   \\ \hline
\end{tabular}
\caption{\small Face recognition performance using super-resolution before face-alignment}
\label{tinyface-superres}
\vspace{-5pt}
\end{table}

\textbf{Baseline:} For the baseline experiment, we choose to follow the usual practice of re-scaling the images to a fixed size irrespective of resolution. We trained our own HR landmark detector (HR-LD) on $20,000$ AFLW images for this purpose. Tinyface gallery and probe images are resized to $128\times128$ and used by the landmark detector as inputs. Using the predicted landmarks, images are aligned to a canonical co-ordinates similar to ArcFace\cite{DBLP:journals/corr/abs-1801-07698}. Baseline performance was obtained by computing cosine similarity between gallery and probe features extracted from the network after feed-forwarding the aligned images.  \vspace{2pt}

\textbf{Setting I1:} \textit{Does adversarial training help?} The model trained for S3 (Section \ref{ablation}) is used to align the images directly in  low resolution. Features for gallery and probe images are extracted after the rescaling the images and cosine distance is used to measure the similarity and retrieve the images from the gallery.   \vspace{2pt}

\textbf{Setting I2:} \textit{Does $G_{2}$ trained in adversarial manner scale to real LR images?} For this experiment, the model trained for L3 in Section \ref{experiments} is used for landmark detection in LR. To recall, in this setting, the three models $G_2$, $D_2$ and $D_3$ (with $G_1$ and $D_1$ frozen) are trained jointly in a semi-supervised way and Tinyface training images are used as real LR data for $D_3$. \vspace{2pt}

\textbf{Setting I3:} \textit{End-to-end training?} In this case, we align the images using the model from setting L4 from Section \ref{experiments}. In this case, we also trained High to low networks ($G_1$ and $D_1$) using training images from Tinyface dataset as real LR images. After training the model for 200 epochs, the weights are frozen to train $G_2, D_2$ and $D_3$ in a semi-supervised way. 

\textbf{\textit{Observation:}} With no surprise, we observe that (from Table \ref{tinyface_ir}) training the heatmap prediction networks in a semi-supervised manner, and aligning the images directly in low resolution, improves the performance of any face recognition system trained with HR images.

\subsection{Additional Experiments:}
\label{additional}
\textbf{Setting A1: }\textit{Does Super-resolution help?} The aim of this experiment is to understand if super-resolution can be used to enhance the image quality before landmark detection. We use SRGAN\cite{DBLP:journals/corr/LedigTHCATTWS16} to super-resolve the images before using face alignment method from Bulat \emph{et al.} \cite{bulat2017far} to align the images.  \vspace{2pt}

\textbf{Setting A2: } \textit{Does Super-resolution help?} In this case, we use ESRGAN\cite{DBLP:journals/corr/abs-1809-00219} to super-resolve the images before using HR-LD (below) to align.

\textbf{\textit{Observation:}} It can be observed from Table \ref{tinyface-superres}, that face recognition  performance obtained after aligning super-resolved images is not at par even with the baseline. It can be hypothesized that possibly super-resolved images do not represent HR images using which \cite{bulat2017far} or HR-LD are trained. \vspace{-5pt}
 
\begin{figure}[h]
    \centering
    \begin{subfigure}[]{ 0.48\textwidth}
        \centering
        \includegraphics[height=2cm, width=\textwidth]{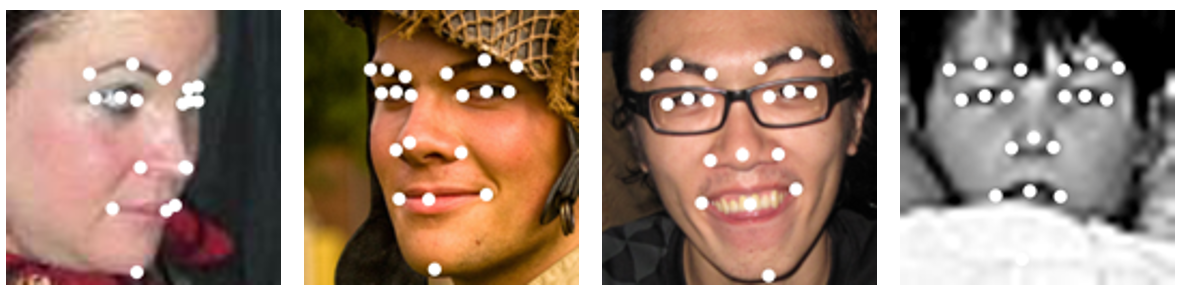}
        \label{fig:aflw_a}
    \end{subfigure}   
    \vspace{-10pt}
    \begin{subfigure}[]{0.48\textwidth}
        \centering
        \includegraphics[height=2cm,width=\textwidth]{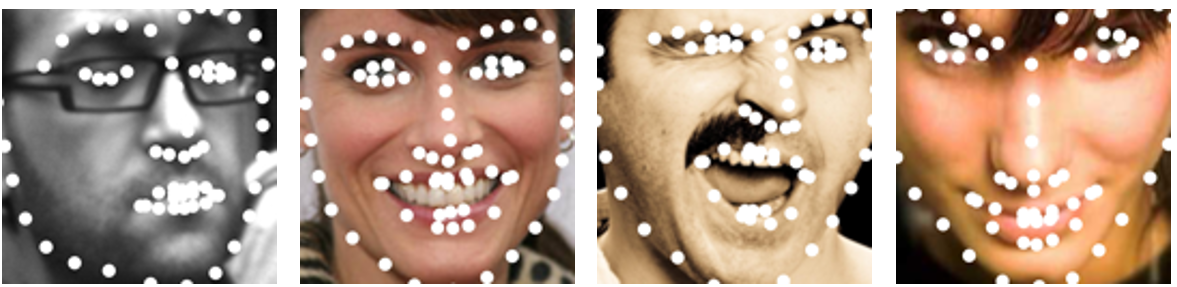}
        \label{fig:300w_a}
    \end{subfigure}
    \vspace{-5pt}
    \caption{\small Sample outputs obtained by training $G_2$ with HR images. First row shows samples from AFLW test set. Second row shows sample images from 300W test set. Last two columns of second row shows outputs from challenging subset of 300W }
     \label{fig:highresalign}
\vspace{-15pt}
\end{figure}

\textbf{High Resolution Landmark Detector (HR-LD)} For this experiment, we train $G_2$ on high resolution images of size $128\times128$ (for AFLW and 300W) using $l_MSE$ loss from Equation \ref{keygen1}. We evaluate the performance of this network on common benchmarks of AFLW-Full test and 300W test sets, shown in Table \ref{ibug_res}. We would like to make a note that LAB\cite{wayne2018lab} and SAN\cite{dong2018san} either uses extra data or extra annotations or larger spatial resolution to train the deep networks. A few sample outputs are shown in Figure \ref{fig:highresalign}
\begin{table}[h]
\centering
\resizebox{\columnwidth}{!}{
\begin{tabular}{|l|l|l|l|l|} 
\hline
\textbf{Method}  &                 & \textbf{300W}               &               & \textbf{AFLW} \\
\hline
  & \textbf{Common} & \textbf{Challenge} & \textbf{Full} & \textbf{Full} \\
\hline \hline
RCPR\cite{10.1109/ICCV.2013.191}             & 6.18                & 17.26             & 8.35   & -                 \\
SDM\cite{XiongD13}              & 5.57                & 15.40             & 7.52         &  5.43          \\
CFAN\cite{CFAN}             & 5.50                & 16.78             & 7.69                & -    \\
LBF\cite{DBLP:conf/cvpr/RenCW014}             & 4.95                & 11.98             & 6.32   &   4.25               \\
CFSS\cite{Zhu_2015_CVPR}             & 4.73                & 9.98              & 5.76         &    3.92        \\
TCDCN\cite{DBLP:conf/eccv/ZhangLLT14}           & 4.80                & 8.60              & 5.54     & -               \\
MDM\cite{trigeorgis2016mnemonic}              & 4.83                & 10.14             & 5.88    & -                \\
PCD-CNN\cite{DBLP:journals/corr/abs-1802-06713}              & 3.67                & 7.62             & 4.44                    & 2.36 \\
SAN\cite{dong2018san}{*}  	& 3.34 	& 6.60 & 3.98 & 1.91 \\ 
LAB\cite{wayne2018lab}{*} & \textbf{2.57} & \textbf{4.72} & \textbf{2.99} & 1.85 \\ \hline \hline
\textbf{HR-LD} & 3.60       & 7.301              & 4.325 & \textbf{1.753} \\         
\hline
\end{tabular}
}
\caption{\small Comparison of the proposed method with other state of the art methods on AFLW (Full) and 300-W testsets. The NMEs for comparison on 300W dataset are taken from the Table 3 of \cite{Lv_2017_CVPR}. In this case $G_2$ is trained in supervised manner using high resolution images of size $128\times128$. {*} uses extra annotation or data.}
\label{ibug_res}
\vspace{-5pt}
\end{table}

\section{Conclusion}
In this paper, we first present an analysis of landmark detection methods when applied to LR images, and the implications on face recognition. We also discuss the proposed method for predicting landmarks directly on LR images. We show that the proposed method improves face recognition performance over commonly used practices of rescaling and super-resolution. As a by-product, we also developed a simple but state of the art landmark detection network. Although, low resolution is chosen as the source of degradation, however, the method can trivially be extended to capture other degradations in the imaging process, such as motion blur or climatic turbulence. In addition, the proposed method can be applied to detect human keypoints in LR in order to improve skeletal action recognition. In the era of deep learning, LR landmark detection and face recognition is a fairly untouched topic, however, we believe this work will open new avenues in this direction.  
\section{Acknowledgment}
This research is based upon work supported by the Office of the Director of National Intelligence (ODNI), Intelligence Advanced Research Projects Activity (IARPA), via IARPA R\&D Contract No. 2014-14071600012. The views and conclusions contained herein are those of the authors and should not be interpreted as necessarily representing the official policies or endorsements, either expressed or implied, of the ODNI, IARPA, or the U.S. Government. The U.S. Government is authorized to reproduce and distribute reprints for Governmental purposes notwithstanding any copyright annotation
thereon.  

{
\bibliographystyle{ieee}
\small
\bibliography{egbib}
}

\end{document}